\pdfoutput=1
\documentclass[11pt]{article}

\usepackage[preprint]{acl}

\usepackage{times}
\usepackage{latexsym}
\usepackage[T1]{fontenc}
\usepackage[utf8]{inputenc}
\usepackage{microtype}
\usepackage{inconsolata}

\usepackage{graphicx}
\usepackage{array}
\usepackage{booktabs}
\usepackage{multirow}
\usepackage{amsmath}
\usepackage{amssymb}
\usepackage{xcolor}
\usepackage{tikz}
\usetikzlibrary{positioning,fit,calc,arrows.meta}

\definecolor{gapcolor}{HTML}{C44E52}
\definecolor{maincolor}{HTML}{3B6EA5}
\newcommand{\gap}[1]{\textcolor{gapcolor}{\underline{#1}}}

\title{Measuring and Reducing Unusable Replies of LLM Agents}

\author{
Yilin Jiang\textsuperscript{1,2},
Fei Tan\textsuperscript{1}\thanks{Corresponding author.},
Xuanyu Yin\textsuperscript{1},
Guanchu Wang\textsuperscript{3},
Xia Hu\textsuperscript{3},
Yutao Yue\textsuperscript{2,4} \\
\textsuperscript{1}East China Normal University, Shanghai, China \\
\textsuperscript{2}The Hong Kong University of Science and Technology (Guangzhou), Guangzhou, China \\
\textsuperscript{3}Shanghai Artificial Intelligence Laboratory, Shanghai, China \\
\textsuperscript{4}Institute of Deep Perception Technology, JITRI, Wuxi, China \\
\texttt{ftan@mail.ecnu.edu.cn}
}

\begin{document}
\maketitle

\begin{abstract}
People delegate multi-step work to LLM agents and read only the reply that the agent hands back. Current evaluation scores task completion and the faithfulness of reports, and it leaves open whether the requester can understand and use the reply. We call this property reply usability and measure it with a reader model that takes the requester's position and tests whether the reply can be interpreted from what the requester has seen, is in the requester's language and answers the request. Its agreement with the majority of five annotators lies within the range of the individual annotators. Across eight models from seven families, 54\% of 480 replies fail, and the best model still fails on 30\%. Failed replies mostly point to material the requester has not seen or use labels that the agent coined during its work. Stating explicit requirements for the reply reduces the failure rate from 0.52 to 0.43 together with errors and omissions, and a secondary model that rewrites the reply reduces it to 0.36, which human annotators confirm in a blind comparison. Reply usability can thus be scored alongside task success and improved with one appended message or one call to a secondary model.

\end{abstract}

\section{Introduction}
\label{sec:intro}

\begin{figure}[t]
\centering
\footnotesize
\newcommand{\mk}[1]{\textsuperscript{\textcolor{gapcolor}{\bfseries #1}}}
\newcommand{\figgutter}{1.4cm}
\newcommand{\figtext}{\dimexpr\columnwidth-1.4cm-13pt\relax}
\begin{tikzpicture}[
  msg/.style={anchor=north west, align=left, inner xsep=0pt, inner ysep=4pt, text width=\figtext},
  lab/.style={anchor=north east, align=right, inner xsep=0pt, inner ysep=4pt, font=\footnotesize\bfseries},
  rail/.style={line width=0.5pt, draw=black!70},
  flow/.style={rail, -{Stealth[length=4.5pt]}}]
\node[msg] (req) at (0,0) {We want a small ($\sim$7B) model to get close to large-model quality on long-document QA, mainly through retrieval and context organization [\ldots] Start by surveying the area and give me a few feasible directions.};
\node[lab] at ([xshift=-7pt]req.north west) {Request};
\node[msg, fill=black!8, inner xsep=5pt, text width=\dimexpr\figtext-10pt\relax] (work) at ([xshift=-5pt,yshift=-3pt]req.south west) {\scriptsize \texttt{search\_papers} $\times$5, \texttt{list\_files}, \texttt{read\_file}\\ \scriptsize \texttt{read\_paper} $\times$10 $\rightarrow$ \textcolor{gapcolor}{\texttt{R07}}, \textcolor{gapcolor}{\texttt{R09}}, \ldots\\ \scriptsize \texttt{run\_study} $\times$22 $\rightarrow$ F1 by domain and budget\\ \scriptsize \texttt{write\_file} $\times$2 $\rightarrow$ notes with \textcolor{gapcolor}{\texttt{reader\_size: 70B}}};
\node[lab, text=black!55] at ([xshift=-2pt]work.north west) {Agent's\\work};
\node[msg] (reply) at ([xshift=5pt,yshift=-3pt]work.south west) {[\ldots] when I passed \gap{reader\_size: 70B}\mk{1} it was silently ignored [\ldots] HyDE is +2.1 on science but $-$2.2 on legal (exactly \gap{R07}\mk{2}'s stated failure mode on exact-match lookups). [\ldots] Clean inverted-U, peak $\sim$8k [\ldots] Replicates and refines \gap{R09}\mk{3} ($\sim$6k).};
\node[lab] at ([xshift=-7pt]reply.north west) {Reply};
\draw[line width=0.8pt] ([xshift=-\figgutter,yshift=1pt]req.north west) -- ([yshift=1pt]req.north east);
\draw[line width=0.8pt] ([xshift=-\figgutter,yshift=-1pt]reply.south west) -- ([yshift=-1pt]reply.south east);
\node[draw=black!45, line width=0.5pt, inner sep=5pt, anchor=north west, text width=\dimexpr\columnwidth-11pt\relax] (frame) at ([xshift=-\figgutter,yshift=-15pt]reply.south west) {%
\renewcommand{\arraystretch}{1.2}%
\begin{tabular*}{\linewidth}{@{}r@{\hspace{5pt}}>{\raggedright\arraybackslash}p{\dimexpr\linewidth-52pt\relax}@{\extracolsep{\fill}}l@{}}
\multicolumn{3}{@{}l}{\textbf{Reader model}}\\[1pt]
\midrule
 & Question & Answer\\
\midrule
\textcolor{gapcolor}{\textbf{1}} & What does ``reader\_size: 70B'' refer to? & none\\
\textcolor{gapcolor}{\textbf{2}} & Which paper [\ldots] does R07 refer to? & none\\
\textcolor{gapcolor}{\textbf{3}} & Which paper [\ldots] does R09 refer to? & none\\
\midrule
\multicolumn{3}{@{}l}{\textcolor{gapcolor}{3 information gaps}}\\
\end{tabular*}};
\coordinate (railtop) at ([xshift=6pt]req.east);
\draw[rail] ([xshift=3pt]req.east) -- (railtop);
\draw[rail] ([xshift=3pt]reply.east) -- (reply.east -| railtop);
\draw[flow] (railtop) -- (railtop |- frame.north);
\end{tikzpicture}
\caption{A reply of Claude Opus 5 on long-document question answering and its reading by the reader model. The requester does not see the agent's work of 41 tool calls. The reader model finds no answer to its questions about the three marked names.}
\label{fig:example}
\end{figure}
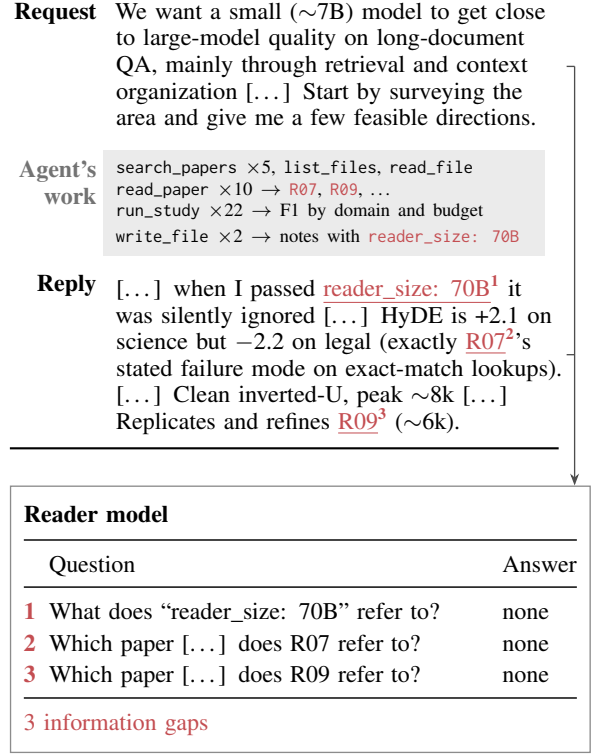

People increasingly hand multi-step work to LLM agents and read only what the agent sends back. An agent that analyzes data, runs experiments or edits a codebase issues dozens of tool calls, writes files and keeps notes, and the requester sees none of this. The message that the agent writes when it returns to the requester, which we call its reply, is the requester's only view of what was done. A requester who cannot read the reply has to recover the result by asking again or by reading the agent's records. Figure~\ref{fig:example} shows such a reply. It reports results under a configuration key and two reference identifiers that the agent met during its work and that the requester has never seen.

Current evaluation does not measure whether the requester can understand and use the reply. Agent benchmarks score whether the task was completed \citep{jimenez2023swe,zhou2024webarena,mialon2024gaia,liu2024agentbench}, and evaluation of generated text scores whether the text is faithful to its source \citep{kryscinski2019evaluating,maynez2020faithfulness}. A reply can satisfy both and still be unusable. It may report the result under a run name that the agent invented, cite a finding that appeared only in a tool output, or list changes without answering the question that was asked. Usability for the requester is therefore a separate property from task success and faithfulness.

We define a reply as failed when the requester cannot interpret it from what they have seen, when it is written in the wrong language, or when it does not answer the request. We measure failures with a reader model that takes the requester's position. The reader sees what the requester has seen, generates the questions that a requester would need answered, and searches for their answers in the reply and in this context (Figure~\ref{fig:example}, bottom). On a gold set labeled by five annotators, the reader agrees with the majority label at $\kappa = 0.60$, within the range that individual annotators reach against the other four.

With this measure we study eight current models from seven families in multi-session collaborations on research and business projects. More than half of the replies fail, 54\% of 480, and every family is affected, with the highest rates for replies that follow extended tool use. The failed replies are mostly written in the terms of the agent's private work. They point to material the requester has not seen or use labels that the agent coined, and more than half of these labels first occur in the agent's own files, project memory or tool-call arguments. Explicit requirements stated at the point of the reply reduce the failure rate from 0.52 to 0.43 together with errors and omissions, whereas a notice about the reader does not. A secondary model that rewrites the reply under the same requirements reduces the rate from 0.52 to 0.36, and annotators who judge default replies and rewrites blind confirm the reduction. Under the requirements, an open-weight model produces a usable reply within sixteen samples for almost every context.

Our contributions are as follows. First, we define the usability of a reply from the requester's position and provide a reader model that measures it at the agreement level of an individual human annotator. Second, we show that unusable replies are common in every model family we evaluate, and we identify what the requester is missing, which is mostly material and vocabulary from the agent's private work. Third, we provide two inexpensive remedies, a requirements message and a rewriting prompt, and we will release the collaborations, the reader and its judgments.

\section{Related Work}
\label{sec:related}

\begin{table}[t]
\centering
\small
\setlength{\tabcolsep}{4pt}
\begin{tabular}{@{}>{\raggedright\arraybackslash}p{2.6cm}>{\raggedright\arraybackslash}p{2.3cm}>{\raggedright\arraybackslash}p{2.2cm}@{}}
\toprule
Line of work & Judged against & Property \\
\midrule
Agent benchmarks & environment state & task success \\
Self-reports of agents & the trajectory & faithfulness, completeness \\
Hand-offs between agents & the next agent's task & downstream success \\
Decontextualization & the empty context & interpretability of a sentence \\
\midrule
This work & what the requester has seen & usability of the reply \\
\bottomrule
\end{tabular}
\caption{What related lines of work evaluate.}
\label{tab:positioning}
\end{table}

\paragraph{Common ground and decontextualization.} Table~\ref{tab:positioning} places our measure among the evaluation targets of related work. Conversation relies on common ground that the participants build together \citep{clark1991grounding}. Speakers are expected to design an utterance for its audience \citep{clark1982audience}, and people do so imperfectly \citep{horton1996speakers,camerer1989curse}. \citet{shaikh2023grounding} show that LLM generations presume common ground and contain fewer grounding acts than human dialogue, and \citet{shaikh2025navigating} measure the resulting failures in logged human--LLM conversations. Decontextualization asks a related question of written text. \citet{choi2021decontextualization} define when a sentence can be interpreted outside its document, and \citet{newman2023question} decontextualize snippets of scientific papers by generating and answering the questions a reader would have \citep[see also][]{gunjal2024molecular}. We adopt this question-answering formulation and change the reference context from the empty context to what the requester has seen.

\paragraph{Reports of agents about their own work.} \citet{bansal2024challenges} name conveying information from the agent to the user as an open challenge, and \citet{shao2024collaborative} include inadequate summaries after executed actions as one category in an error taxonomy of human--agent collaboration. \citet{tang2026coding} and \citet{kraishan2026plans} find that self-reports of coding agents misstate or leave out executed actions, and \citet{huang2026models} study whether agents disclose problems in their work. These studies judge the report against the trajectory, with annotators or checkers who have seen the trajectory, and they measure faithfulness and completeness. We measure a complementary property, whether a requester who has not seen the trajectory can understand and use the report, along with faithfulness. A further line of work studies hand-offs of coding tasks between agents and their effect on quality and cost \citep{ganz2026handoff,kc2026handoff}, misalignment between agents in multi-agent systems \citep{cemri2025multi}, and the loss of earlier constraints in context compaction \citep{chen2026governance}. In our setting the reader is a person, and the outcome is whether that person can read the message from what they have seen.

\paragraph{Instructions about the audience.} Instructions that tell a model about its audience have shown limited effect on other tasks, as have personas in system prompts \citep{zheng2024helpful}. \citet{shaikh2023grounding} report that prompting for grounding acts raises their frequency and yields at most minimal gains in agreement with human behavior. \citet{christian2026self} find that instructing a model to ignore information it has seen does not remove the influence of that information on its judgments. \citet{wang2026unreliable} report that reminders recover little of the error in the progress reports of agents, and \citet{chen2026governance} reports that provenance prompts in the style of Spotlighting do not reduce constraint violations after context compaction, whereas pinning the constraints does. We compare a notice about the reader with a concrete specification of what the reply must contain, in the same context and with the same information (Section~\ref{sec:reducing}).

\paragraph{Readability and faithfulness.} Rewriting text for a reader can change its content. Lay summarization \citep{goldsack2022making} and personalized jargon identification \citep{guo2024personalized} address readers outside a text's field, and \citet{devaraj2022evaluating} show that text simplification introduces factual errors and omissions. We therefore check the replies of every writing condition and every rewrite against the work record. Other work changes the context in which a text is written. \citet{zhu2026fs} report gains in report quality, scored against references, from separating the writing of a research report from the search that precedes it, and \citet{laban2025get} and \citet{chen2026self} show for task performance and for error correction that the form in which information is presented to a model matters when the information itself is held fixed. For the replies of agents, our writing conditions separately vary the context in which the reply is written and the requirements that it is given.

\section{Measuring Reply Usability}
\label{sec:measurement}

\subsection{Reply Usability}
\label{sec:setting}

We study the message an agent sends back to the person who asked for the work. In a session, a requester gives an instruction, the agent works through a sequence of tool calls, and the agent then writes a message without tool calls. We call this message the agent's \emph{reply}. The reply is often the only part of the session the requester reads, so the requester has to be able to read it without consulting anything else.

The requester and the agent see different parts of the session. The \emph{requester context} $R$ contains the requester's own messages, the files the requester provided, and the agent's earlier replies. The agent also sees its \emph{private context}, which we define as everything outside $R$ that the agent produced or received during the work. It comprises tool calls and tool outputs, documents and notes in the agent's workspace, summaries written during context compaction, and memory loaded at the start of a new session. The requester has seen none of the private context.

We call a reply \emph{usable} when the requester can understand it from $R$ alone and it answers what was asked, which we specify with three criteria. First, the reply can be interpreted from $R$. We adapt the definition of decontextualization by \citet{choi2021decontextualization}, who call a sentence decontextualized when it can be interpreted in an empty context with its truth-conditional meaning preserved, and we replace the empty context with $R$. An \emph{information gap} is an expression of the reply whose interpretation requires information that neither the reply nor $R$ provides, such as a run name, a configuration label, or a finding that appeared only in a tool output. Such an expression presupposes common ground \citep[in the sense of][]{clark1991grounding,shaikh2023grounding} that the requester lacks, and a reply meets the first criterion when it has no information gap. Second, the reply is written in the requester's language. Third, the reply answers the request. A reply that lists the changes the agent made and does not state the result the requester asked for gives less information than the exchange requires \citep{grice1975logic}. A reply \emph{fails} when it violates at least one criterion, and the \emph{failure rate} of a set of replies is the proportion that fail.

Usability in this sense is distinct from task success and from faithfulness. An agent can complete the task and describe its work accurately in a reply that the requester cannot interpret, and a reply that the requester can read with ease can still misreport the work. We therefore measure faithfulness to the work alongside usability (Section~\ref{sec:faithfulness}).

\subsection{Measuring Usability with a Reader Model}
\label{sec:reader}

We measure failures with a \emph{reader model} that takes the position of the requester. The reader knows the requester's field, the request and the requester's files, it has read $R$, and it has not seen the agent's work. It reads at an ordinary pace and accepts only what the text states.

The reader follows question-answering formulations of decontextualization \citep{newman2023question} and of summary consistency \citep{wang2020asking,durmus2020feqa,scialom2021questeval} and runs in two steps. In the first step it generates the questions that the requester would need answered to understand the reply. Each question is tied to a span of the reply and typed as a definition, a reference or a background question. In the second step the reader looks for the answer to each question in the text around the span, then in the whole reply, and then in $R$, and it must quote the passage that answers the question. A question without a quoted answer marks an information gap. The first step also returns the sentence of the reply that answers the requester's latest message, and the language of the reply is determined from its characters. A reply fails if it has at least one information gap, is written in a language other than the requester's, or contains no sentence that answers the request.

The reader uses Gemini 3 Flash (preview). Comparisons between writing conditions and between a reply and its rewrite are paired within the point of the collaboration at which the reply is written. We report 95\% percentile intervals from a bootstrap that resamples collaborations. Appendix~\ref{app:reader} gives the prompts and settings.

\begin{figure}[t]
\centering
\includegraphics[width=\columnwidth]{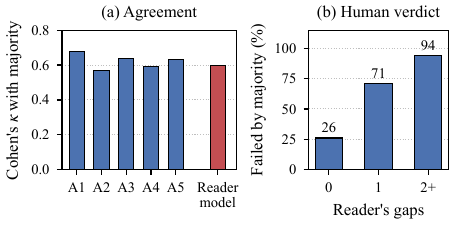}
\caption{(a) Cohen's $\kappa$ with the majority label of the gold set, which for an annotator is the majority of the other four. (b) Replies failed by the majority, by the number of gaps the reader finds.}
\label{fig:validation}
\end{figure}

\paragraph{Agreement with human annotators.}\label{sec:agreement} The reader's agreement with the human majority lies within the range of the individual annotators (Figure~\ref{fig:validation}a). We validated the reader on a gold set of 300 replies. Five annotators with research backgrounds in five fields labeled every reply. The annotators read each collaboration from its beginning, saw only $R$, labeled each reply as readable or failed, and marked the spans that stopped them. The annotators agree with each other at Fleiss' $\kappa = 0.63$, and the reader agrees with their majority label at Cohen's $\kappa = 0.60$ with an accuracy of 82\%.

The reader's verdict tracks what the annotators experience. The share of replies that the majority fails rises with the number of information gaps the reader finds, from 26\% for replies without a gap to 71\% for replies with one gap and 94\% for replies with two or more (Figure~\ref{fig:validation}b), so the majority already fails most replies with a single gap. Of the gaps the reader finds, 68\% lie on a span that at least one annotator marked. Annotators also take longer over the replies that the majority fails, with a median reading time of 85 seconds against 50 seconds for those it passes (Appendix~\ref{app:agreement}).

\subsection{Measuring Faithfulness with a Checker Model}
\label{sec:faithfulness}

We check faithfulness to the work with a \emph{checker model}. The checker receives the reply and the \emph{work record}, which is the agent's context at the point of the reply, including its tool calls and tool outputs. It lists the factual errors, which are statements of the reply that the record contradicts or does not contain, and the omissions, which are key results of the work that the reply leaves out. We report the number of errors plus omissions per reply. The checker uses the same model as the reader, and Appendix~\ref{app:reader} gives its prompt.

\section{Experimental Setup}
\label{sec:setup}

\subsection{Collaborations}
\label{sec:collaborations}

We study replies in multi-session collaborations in which an agent works for a scripted requester. The requester gives brief directives and asks conceptual questions, as a researcher or an analyst does when delegating work. After each requester message the agent calls tools freely, and its first message without tool calls is its reply. A collaboration has 15 requester messages in three sessions. When the context exceeds a length threshold, the history is replaced by a summary that the agent writes, and each new session starts from the documents in the agent's workspace, which are loaded as project memory.

\begin{table}[t]
\centering
\small
\begin{tabular}{@{}>{\raggedright\arraybackslash}p{4.5cm}ll@{}}
\toprule
Project & Type & Language \\
\midrule
Detecting mislabeled examples in crowd-labeled datasets & research & Chinese \\
Long-document question answering with a small model and retrieval & research & English \\
Price change for a subscription product & business & Chinese \\
Redesign of the onboarding flow of a software product & business & English \\
\bottomrule
\end{tabular}
\caption{The four projects of the collaborations.}
\label{tab:setup}
\end{table}

The collaborations cover four projects, two in research and two in business (Table~\ref{tab:setup}). In the research projects the agent searches and reads a literature index, runs simulated studies whose design it specifies, and interprets the results. In the business projects it consults a reference library and evaluates designs with a simulator. The system prompt is a generic agent instruction and does not mention who reads the replies. Because the environments are scripted, every model works on the same projects and receives the same requests. Appendix~\ref{app:collaborations} gives the tools, the system prompt and the limits.

\subsection{Models}
\label{sec:models}

We evaluate eight current models from seven families. The set contains Claude Opus 5, GPT-5.6 Sol, DeepSeek V4.1 Flash, DeepSeek V4 Pro, Kimi K2.6, GLM-5.3, Qwen3.6 Plus and MiniMax M3, and it includes closed and open-weight models. Each model works on the four projects, which gives 32 collaborations and 480 replies.

\section{Replies Fail Across Model Families}
\label{sec:prevalence}

\begin{figure*}[t]
\centering
\includegraphics[width=\textwidth]{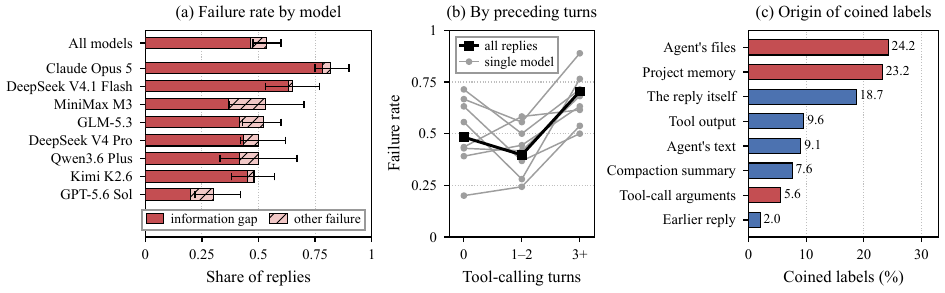}
\caption{(a) Failure rate of each model with 95\% intervals, divided into replies with an information gap and other failures. (b) Failure rate by tool-calling turns before the reply. (c) First occurrence of the coined labels in the agent's context, where red marks the agent's files, project memory and tool-call arguments.}
\label{fig:failures}
\end{figure*}

\subsection{Prevalence}
\label{sec:prevalence-rates}

More than half of the replies fail, and the failure appears in every family. Figure~\ref{fig:failures}a reports the failure rate of each model when it works with the generic system prompt and no instruction about its replies. Of the 480 replies, 258 fail, a rate of 0.54. Seven of the eight models fail on at least 48\% of their replies, and the rate ranges from 0.30 for GPT-5.6 Sol to 0.82 for Claude Opus 5. By project, the rate ranges from 0.41 for subscription pricing to 0.66 for the detection of mislabeled examples (Appendix~\ref{app:collaborations}).

All seven families are affected, closed and open-weight models alike, and the highest rate belongs to a closed frontier model. The system prompt asks for no particular style of reply, so these rates describe what a requester receives by default.

Most failures are information gaps. Of the 258 failed replies, 222 contain at least one question that the requester cannot answer from $R$, 73 do not answer the request, and 17 are written in the wrong language, where a reply can meet more than one of these conditions. At least one information gap occurs in 46\% of all replies and in 20\% to 78\% of the replies of a single model.

Gaps are frequent within a failed reply and occur early. Of the 258 failed replies, 114 contain two or more information gaps, and a reply with a gap contains 2.1 gaps on average and up to ten. In 128 of the 222 replies with a gap, the first gap begins in the first third of the reply, so the requester is stopped before reaching most of the content.

Failures concentrate in the replies that report the most work (Figure~\ref{fig:failures}b). Replies written after three or more tool-calling turns fail at a rate of 0.70, against 0.48 for replies written without a tool call and 0.40 for replies after one or two turns. In six of the eight models, replies after three or more turns fail more often than replies without a tool call. The two models with the highest failure rates, Claude Opus 5 and DeepSeek V4.1 Flash, are also the two with the most tool-calling turns before a reply (Appendix~\ref{app:collaborations}). The replies that carry the results of extended autonomous work are thus the ones a requester is least likely to be able to use.

\subsection{What the Requester Is Missing}
\label{sec:gaps}

\begin{table*}[t]
\centering
\small
\begin{tabular}{@{}l>{\raggedright\arraybackslash}p{4.25cm}>{\raggedright\arraybackslash}p{4.9cm}r@{}}
\toprule
Kind of failure & What the requester lacks & Example span and the reader's question & Failed replies (\%) \\
\midrule
Pointer to unseen material & a file section, an index entry or the output of a run that the requester has not seen & \texttt{R03}. Which paper or reference does R03 refer to? & 46.1 \\
\addlinespace[2pt]
Coined label & the meaning of a name, number or short form that the agent introduced during the work & \texttt{crn\_hierbm25\_L}. What does ``crn\_hierbm25\_L'' refer to? & 42.2 \\
\addlinespace[2pt]
Unexplained term & the meaning of an abbreviation or term that is established in the field & CUPED adjustment. What does CUPED stand for in the context of variance reduction in A/B testing? & 23.6 \\
\addlinespace[2pt]
Missing background & a premise that the reply depends on and does not state & confirm whether the difference between 80.2 and 76.0 is stable. At what sample size [\ldots] was the comparison of 80.2 and 76.0 run? & 1.6 \\
\midrule
Request not answered & \multicolumn{2}{@{}l}{\hspace{\tabcolsep}the result or answer that the latest message asked for} & 28.3 \\
Wrong language & \multicolumn{2}{@{}l}{\hspace{\tabcolsep}a reply in the requester's language} & 6.6 \\
\bottomrule
\end{tabular}
\caption{Kinds of failure among the 258 failed replies, where a reply can have several kinds. The upper four are the categories of information gaps, shown with a span of a reply and the reader's unanswered question.}
\label{tab:kinds}
\end{table*}

Most failed replies rely on material that exists only in the agent's private context. For each failed reply the reader returns the questions it could not answer, and each marks a gap. The 258 failed replies contain 475 gaps, which a classifier assigns to four categories, and two further kinds of failure come from the reader's verdict (Table~\ref{tab:kinds}). On 100 gaps that two annotators labeled independently, the classifier agrees with the annotators at a mean $\kappa$ of 0.68, the level at which the two annotators agree with each other (Appendix~\ref{app:gaps}).

Pointers to unseen material and coined labels each occur in more than 40\% of the failed replies. In six of the seven families, at least a third of the failed replies contain a pointer to unseen material and at least a quarter contain a coined label. For Claude Opus 5, the model with the highest failure rate, the two shares are 61.2\% and 65.3\%. GPT-5.6 Sol, the model with the lowest rate, has a different profile. None of its 18 failed replies contains a coined label, 44.4\% leave an established term unexplained and 38.9\% do not answer the request. The work itself therefore does not force working labels into the reply.

More than half of the coined labels were written down by the agent before it wrote the reply. For each coined label we search the agent's context for the earliest message that contains it (Figure~\ref{fig:failures}c). Of the labels, 24.2\% first occur in a file that the agent wrote and 5.6\% in the arguments of another tool call. A further 23.2\% first occur in the project memory loaded at the start of a later session, which consists of documents that the agent wrote in earlier sessions. These three sources account for 53.0\% of the labels.

These origins show that the vocabulary of a failed reply is the working vocabulary of the agent. During the work the agent names its runs, configurations and intermediate findings, and it records the names in its notes and memory. In the reply it continues to use these names and cites its sources by the identifiers under which it retrieved them, none of which the requester can resolve from $R$. Project memory extends this vocabulary beyond the session in which it arose, since 23\% of the coined labels reach the reply from documents that the agent wrote in an earlier session. In the reply of Figure~\ref{fig:example}, the key \texttt{reader\_size: 70B} occurs in the agent's notes, and \texttt{R07} and \texttt{R09} are the identifiers under which the literature index returned two papers.

\section{Reducing Failures}
\label{sec:reducing}

\begin{table*}[t]
\centering
\small
\setlength{\tabcolsep}{5pt}
\begin{tabular}{@{}lllccc@{}}
\toprule
Condition & Written in & Added instruction & Failure rate & Reduction & \shortstack{Errors +\\omissions} \\
\midrule
Default & working context & none & 0.52 & & 1.41 \\
Notice & working context & notice & 0.50 & 0.027 [$-$0.052, 0.112] & 1.34 \\
Requirements & working context & requirements & \textbf{0.43} & \textbf{0.098} [0.030, 0.168] & \textbf{1.07} \\
\midrule
Separate pass & new conversation & none & 0.47 & 0.051 [$-$0.004, 0.109] & 1.37 \\
Fact list & new conversation & none & 0.49 & 0.031 [$-$0.020, 0.081] & 1.45 \\
Separate pass with requirements & new conversation & requirements & \textbf{0.42} & \textbf{0.102} [0.023, 0.181] & 1.16 \\
\bottomrule
\end{tabular}
\caption{The six writing conditions on 256 reply points, with 95\% intervals for the reduction from the default.}
\label{tab:conditions}
\end{table*}

\subsection{Writing Conditions}
\label{sec:conditions}

We test interventions at 256 reply points of the collaborations of Section~\ref{sec:prevalence}. At each point the reply is written again under every condition while the work up to that point stays fixed.

Three conditions write the reply in the agent's working context. The default condition adds no instruction. The notice condition adds to the system prompt a statement that the user has seen only their own messages and the agent's earlier replies. The requirements condition appends a message at the point of the reply that repeats this statement and specifies what the reply must contain: an answer to the latest message first, what was done, the result, the next step, where the evidence can be checked, and a definition of every name the user has not seen. These items correspond to the kinds of failure of Section~\ref{sec:gaps}. The answer to the latest message addresses unanswered requests, and the definitions address coined labels and unexplained terms.

Three conditions write the reply in a separate pass in a new conversation. In the first the model receives $R$ and the work record. In the second it receives $R$ and a list of the results that it extracted from its own work. In the third it receives $R$, the work record and the requirements. Appendix~\ref{app:conditions} lists the prompts.

\subsection{Explicit Requirements}
\label{sec:requirements}

Explicit requirements reduce the failure rate from 0.52 to 0.43, a reduction of 0.098 (Table~\ref{tab:conditions}). The rate is lower in seven of the eight models and in six of the seven families (Figure~\ref{fig:reducing}a). The reduction is largest for Claude Opus 5, the model with the highest default rate, whose failure rate falls from 0.72 to 0.53, and four further models improve by 0.12 or more.

The requirements also improve faithfulness. The checker lists 1.07 errors plus omissions per reply under the requirements against 1.41 by default. Both components decrease, errors from 0.73 to 0.53 per reply and omissions from 0.68 to 0.54.
A notice about the reader has no such effect, with a failure rate of 0.50 and 1.34 errors plus omissions.

The reduction follows the requirements message, and the message works in the agent's own context. The two conditions that include it bring the failure rate to 0.43 and 0.42, and none of the three conditions without it differs significantly from the default (Table~\ref{tab:conditions}). The difference between these two conditions is 0.004 [$-$0.069, 0.075], so a separate writing step adds no measurable reduction to one appended message, which needs no additional call.

\subsection{Rewriting by a Secondary Model}
\label{sec:rewriting}

\begin{figure*}[t]
\centering
\includegraphics[width=\textwidth]{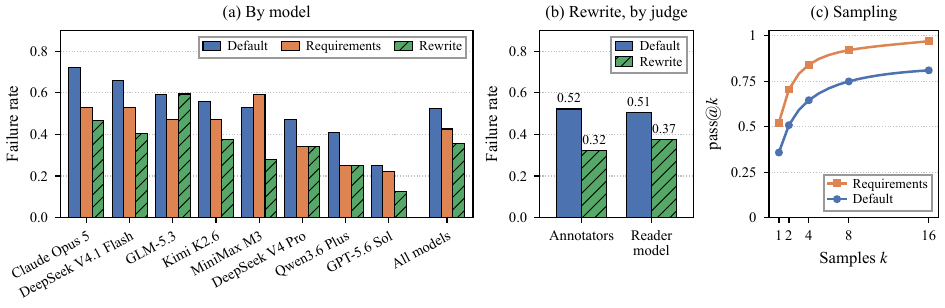}
\caption{(a) Failure rate by default, under the requirements and after the rewrite. (b) Default replies and rewrites at 83 reply points, judged by annotators and by the reader. (c) Share of reply points with a usable reply among $k$ samples.}
\label{fig:reducing}
\end{figure*}

A secondary model can repair a reply after it has been written. This setting applies when the agent's prompt cannot be changed or the reply already exists. The rewriting model is DeepSeek V4 Pro. It receives the request, the work record and the agent's default reply, and it rewrites the reply once under the requirements of Section~\ref{sec:conditions}. We compare each rewrite with the reply it replaces at the same 256 reply points.

Rewriting reduces the failure rate from 0.52 for the default replies to 0.36 for their rewrites, a reduction of 0.168 [0.101, 0.242]. The rate is lower in seven of the eight models and in six of the seven families (Figure~\ref{fig:reducing}a). The rewrite also reduces failures in the replies of the two closed models, from 0.72 to 0.47 for Claude Opus 5 and from 0.25 to 0.13 for GPT-5.6 Sol.
Human annotators confirm the reduction. Five annotators labeled the default reply and its rewrite at 83 reply points under the protocol of Section~\ref{sec:agreement}, each seeing one version per point without knowing which. The share of annotators who fail a reply falls from 0.52 for the default reply to 0.32 for the rewrite, a difference of 0.201 [0.095, 0.321] (Figure~\ref{fig:reducing}b), and it is lower in seven of the eight models. On the 166 replies of this evaluation the reader agrees with the majority of annotators at $\kappa = 0.60$ (Appendix~\ref{app:humanrewrite}).

Rewriting also reduces errors and omissions. The checker lists 1.41 errors plus omissions per default reply and 1.13 per rewrite, with errors going from 0.73 to 0.54 and omissions from 0.68 to 0.59. The rewriting model has the work record, so it can define a name or replace a pointer with the content it points to. In one reply of Claude Opus 5, for example, the default text states that ``\texttt{read\_file} truncates at $\sim$8k characters'', and the rewrite states that ``the file-viewing tool truncates long files at roughly 8k characters''. The rewriting model is open-weight, so a model that can be run locally repairs the replies of both closed models.

\subsection{Sampling}
\label{sec:sampling}

Usable replies are within the model's own output distribution. We measure this on Qwen3.6-35B-A3B, an open-weight model run as the agent in further collaborations of the same kind. For 100 reply points we draw 16 samples under the default condition and 16 under the separate pass with requirements. A sample is usable when the reader accepts it in two independent judgments, and pass@$k$ is the share of points with a usable reply among $k$ samples (Appendix~\ref{app:sampling}).

Under the requirements, pass@$k$ is 0.52 for one sample, 0.84 for four and 0.97 for sixteen (Figure~\ref{fig:reducing}c). Under the default condition the values are 0.36, 0.65 and 0.81, so four samples under the requirements cover more points than sixteen samples by default. The model can therefore write a usable reply for almost every context, and selecting among a few samples is a direct route to further reduction.

\section{Discussion}
\label{sec:discussion}

The results locate the failure at the point where an agent turns from working to reporting. During the work the agent builds a vocabulary of run names, configuration labels and identifiers of sources, and its notes and memory keep this vocabulary available. The reply is written from inside that state, so a requester who has seen none of it meets the names without their referents. A notice that the requester has not seen the work does not measurably lower the failure rate, whereas a statement of what the reply must contain does. The models can therefore write for the requester, and a notice of who reads does not by itself lead them to do so. Two consequences follow for the field. An evaluation that inspects only the final state of the environment credits work whose report the requester cannot use, and a reader model that takes the requester's position supplies the missing check. In the design of agents, memory and compaction carry working names across sessions and into later replies, so the moment at which the agent returns to the requester deserves an instruction of its own, and a secondary model can enforce it where the agent's prompt is fixed. Reply usability also bears on how LLMs shape human language. Widespread use of LLMs narrows the linguistic diversity of human writing \citep{sourati2026shrinking}, and the vocabulary that agents carry into their replies is a further possible route for this influence.

\section{Conclusion}
\label{sec:conclusion}

The reply is the requester's only view of an agent's work, and current evaluation does not test whether the requester can understand and use it. We defined usability from the requester's position and built a reader model that measures it at the agreement level of human annotators. Across eight models from seven families, more than half of the replies fail, mostly because they are written in the terms of the agent's private work, and explicit requirements or a rewrite by a secondary model reduce the failures, which human annotators confirm for the rewrite. Reply usability can therefore be measured alongside task success and improved with one appended message or one call to a secondary model.

\section*{Limitations}

Our collaborations are scripted, which gives every model the same requests and records the agent's private context that the analysis of gap origins requires. The reader represents a requester who knows the field of the request and has not seen the agent's work, and the gold set validates it for this profile. The sampling analysis covers one open-weight model. Applying the measure to logged sessions of consenting users, to requesters with other backgrounds, and to further models and languages is the direct next step.

\section*{Ethics Statement}

All collaborations are synthetic and contain no personal data. The annotators of the gold set, of the evaluation of rewrites and of the gap categories are researchers who took part voluntarily, gave informed consent to the use of their labels, and received fair compensation. The annotations are used to validate the reader and the gap classifier and to evaluate the rewrites, and they are not released.

\bibliography{references}

\appendix
\section{Reader and Checker}
\label{app:reader}

The reader is an LLM-based evaluator \citep{liu2023geval,zheng2023judging}. It runs in two steps with Gemini 3 Flash (preview), with thinking disabled, temperature 0 and at most 3,000 output tokens per call. The first step receives the requester context and the reply and returns the questions that a requester would need answered, each with the span of the reply it concerns and a type (definition, reference or background). Questions that readers ask to fill gaps in their understanding of a text are studied as inquisitive questions \citep{ko2020inquisitive}. Identifiers that occur in the reply and in the agent's private context and nowhere in the requester context are added as reference questions by lexical matching. Questions whose span is a file name or a path are removed, since the requester can open that location. The second step looks for the answer to each question in the text around the span, then in the whole reply, and then in the requester context, and must quote the passage that answers it. A question without a quoted answer marks an information gap. The first step also returns whether the requester's latest message asks for something and, if so, the sentence of the reply that answers it. Whether the reply is in the requester's language is determined from the scripts of its characters.
The checker receives the reply and the work record. The work record is the agent's context at the point of the reply after the system prompt, limited to its last 120,000 characters when it is longer. The checker also returns a list of downplayed problems, which the reported count does not include. Intervals in this paper are percentile intervals of a bootstrap \citep{efron1994introduction} over collaborations. The prompts follow as the models receive them, with the fields in braces filled with the requester's files, the conversation, the reply and the work record. Collaborations in Chinese use the Chinese versions of these prompts.

\smallskip\noindent\textbf{Reader, question generation.}\par
{\ttfamily\scriptsize\raggedright\setlength{\parindent}{0pt}\setlength{\parskip}{0pt}\par
You are the person who wrote these requests. You have professional background in the field your requests are about and know its common terms and methods; you know your own requests and the files you provided, but you have not seen the assistant\textquotesingle{}s work process (the tools it called, the outputs it saw, the documents and notes it wrote).\par
\vspace{0.6\baselineskip}
[Files you provided]\par
\{files\}\par
\vspace{0.6\baselineskip}
[Conversation you have seen]\par
\{dialog\}\par
\vspace{0.6\baselineskip}
[Assistant\textquotesingle{}s latest message]\par
\{reply\}\par
\vspace{0.6\baselineskip}
Read the latest message at your normal pace. List the questions you would need answered before you could understand it, each with a type:\par
- \textquotedbl{}definition\textquotedbl{}: what a term or abbreviation means (common terms of your field do not count);\par
- \textquotedbl{}reference\textquotedbl{}: which specific thing an expression refers to (e.g. a code, an ID, a name the assistant made up, \textquotedbl{}that option\textquotedbl{});\par
- \textquotedbl{}background\textquotedbl{}: what background you would need in order to understand it.\par
Each question\textquotesingle{}s \textquotedbl{}about\textquotedbl{} must be a short span copied verbatim from the latest message. Do not ask about things the message explains where it uses them. List only questions without which you cannot understand the message; if none, give an empty list.\par
Also decide whether your latest message asks the assistant for something (a result, conclusion, recommendation, ideas, an outline, an answer...) (answer\_needed: true/false). If it does, copy verbatim the one sentence of the latest message that best answers it; if the message does not give what you asked for (e.g. it only lists actions or changes made), write \textquotedbl{}NONE\textquotedbl{}.\par
\vspace{0.6\baselineskip}
Ask only questions of the kind \textquotedbl{}what does this mean\textquotedbl{}, \textquotedbl{}which thing does this refer to\textquotedbl{}, \textquotedbl{}what background is needed to understand this\textquotedbl{}, without which you cannot understand the message; do not ask for more detail (e.g. \textquotedbl{}how exactly is it computed\textquotedbl{}, \textquotedbl{}what exactly does it include\textquotedbl{}, \textquotedbl{}how exactly will it be done\textquotedbl{}).\par
Output only JSON: \{\textquotedbl{}questions\textquotedbl{}: [\{\textquotedbl{}q\textquotedbl{}: \textquotedbl{}question\textquotedbl{}, \textquotedbl{}type\textquotedbl{}: \textquotedbl{}definition/reference/background\textquotedbl{}, \textquotedbl{}about\textquotedbl{}: \textquotedbl{}verbatim span\textquotedbl{}\}], \textquotedbl{}answer\_needed\textquotedbl{}: true or false, \textquotedbl{}answer\_quote\textquotedbl{}: \textquotedbl{}verbatim or NONE\textquotedbl{}\}\par
}

\smallskip\noindent\textbf{Reader, answer location.}\par
{\ttfamily\scriptsize\raggedright\setlength{\parindent}{0pt}\setlength{\parskip}{0pt}\par
Below is a text (a user\textquotesingle{}s files, their conversation with an assistant, and the assistant\textquotesingle{}s latest message) and some questions. Answer each question using ONLY this text, not your own knowledge, attaching one verbatim sentence from the text that supports the answer. If the text does not contain the answer, answer \textquotedbl{}NOT\_IN\_TEXT\textquotedbl{}.\par
\vspace{0.6\baselineskip}
[Text]\par
\{text\}\par
\vspace{0.6\baselineskip}
[Questions]\par
\{qs\}\par
\vspace{0.6\baselineskip}
Output only a JSON list in the same order as the questions: [\{\textquotedbl{}answer\textquotedbl{}: \textquotedbl{}... or NOT\_IN\_TEXT\textquotedbl{}, \textquotedbl{}quote\textquotedbl{}: \textquotedbl{}verbatim\textquotedbl{}\}]\par
}

\smallskip\noindent\textbf{Faithfulness checker.}\par
{\ttfamily\scriptsize\raggedright\setlength{\parindent}{0pt}\setlength{\parskip}{0pt}\par
You check a message an AI agent sent to its user against the agent\textquotesingle{}s full work record.\par
\vspace{0.6\baselineskip}
[Work record]\par
\{traj\}\par
\vspace{0.6\baselineskip}
[Message to the user]\par
\{reply\}\par
\vspace{0.6\baselineskip}
List (a) factual errors: statements in the message contradicted by or absent from the record; (b) omissions: key results in the record the user needed but the message leaves out; (c) downplayed problems: failures, caveats or negative results that the message hides or softens. Output only JSON: \{\textquotedbl{}errors\textquotedbl{}: [...], \textquotedbl{}omissions\textquotedbl{}: [...], \textquotedbl{}downplayed\textquotedbl{}: [...]\}.\par
}

\section{Agreement with Annotators}
\label{app:agreement}

The research backgrounds of the five annotators cover artificial intelligence, educational technology, intelligent transportation, human--computer interaction and financial technology. Annotators labeled each of the 300 replies of the gold set after reading the collaboration from its beginning in the requester's view. They saw the requester's files, the conversation so far and the current reply, and they did not see tool calls, tool outputs or documents written by the agent. They did not know which model wrote a reply. They read each reply once at their normal pace, could scroll back in the conversation, and did not consult external sources.

For each reply the annotators chose one of three labels modeled on the categories of \citet{choi2021decontextualization}, with the third label extended to replies that do not answer the question. The reply can be read as it is, it can be read after some information is added, such as a definition, the referent of an expression or background, or it cannot be repaired or does not answer the question. The last two labels count as failed. For failed replies the annotators also marked the spans that stopped them and assigned each span one of three types, a missing definition, an unresolved reference or missing background. Annotation began after two practice collaborations and one pilot collaboration. The gold label of a reply is the majority label. Table~\ref{tab:agreement} reports the agreement as Cohen's $\kappa$ \citep{cohen1960coefficient}, and agreement among the five annotators is Fleiss' $\kappa$ \citep{fleiss1971measuring}.

The majority of annotators fail 191 of the 300 replies, 115 with the second label and 76 with the third. The reader fails 171 of these 191 replies and 34 of the 109 replies that the majority passes. Table~\ref{tab:goldsteps} relates the two sets of judgments in both directions. The upper part gives the share of replies that the majority fails by the number of information gaps the reader finds, together with the mean number of annotators who fail the reply. The lower part gives the reader's failure rate by the number of annotators who fail the reply. In the replies they fail, annotators mark a mean of 2.1 spans, and they mark at least two spans in 62\% of them. Of the reader's gaps, 68\% lie on a span marked by at least one annotator, and the reader finds 61\% of the spans marked by the majority. The reader's verdict that a reply does not answer the request agrees with the annotators' majority judgment that the reply states no result on 290 of the 300 replies. The median reading time is 85 seconds for replies that the majority fails and 50 seconds for replies that it passes.

\begin{table}[t]
\centering
\small
\setlength{\tabcolsep}{5pt}
\begin{tabular}{@{}lrcc@{}}
\toprule
Reader's gaps & Replies & \shortstack{Failed by\\the majority} & \shortstack{Annotators\\who fail} \\
\midrule
0 & 108 & 26\% & 1.4 \\
1 & 75 & 71\% & 3.5 \\
2 or more & 117 & 94\% & 4.6 \\
\midrule
Annotators who fail & Replies & \multicolumn{2}{c}{Failed by the reader} \\
\midrule
0 & 57 & \multicolumn{2}{c}{19\%} \\
1 & 28 & \multicolumn{2}{c}{39\%} \\
2 & 24 & \multicolumn{2}{c}{50\%} \\
3 & 22 & \multicolumn{2}{c}{73\%} \\
4 & 30 & \multicolumn{2}{c}{83\%} \\
5 & 139 & \multicolumn{2}{c}{94\%} \\
\bottomrule
\end{tabular}
\caption{Human judgments by the number of gaps the reader finds, and the reader's failure rate by the number of annotators who fail the reply.}
\label{tab:goldsteps}
\end{table}

\begin{table}[t]
\centering
\small
\begin{tabular}{lcc}
\toprule
Rater & $\kappa$ & Accuracy \\
\midrule
Annotator 1 & 0.68 & -- \\
Annotator 2 & 0.57 & -- \\
Annotator 3 & 0.64 & -- \\
Annotator 4 & 0.59 & -- \\
Annotator 5 & 0.63 & -- \\
Reader & 0.60 & 82\% \\
\bottomrule
\end{tabular}
\caption{Agreement with the majority label of the gold set (Cohen's $\kappa$). For each annotator the majority is taken over the other four.}
\label{tab:agreement}
\end{table}

\section{Collaborations}
\label{app:collaborations}

Each environment gives the agent a workspace with tools to list, read and write files, and a set of task tools. The agent calls them in a loop of reasoning and acting \citep{yao2023react}. In the research projects the agent searches a literature index, reads the entries it retrieves, and runs simulated studies whose design it specifies. In the business projects the agent searches and reads a reference library and runs a simulator on designs it specifies. Each environment computes its outcomes from a fixed response surface with interactions between design choices, so the results are reproducible. The agent may take up to 30 tool-calling turns per requester message. When the limit is reached, the agent is asked to reply and may take up to four further turns, and its first message with text is taken as the reply.

The system prompt reads as follows.
\begin{quote}\small
You are an autonomous research assistant working in a sandboxed workspace. Use the tools to carry out the task step by step. Keep a running notes file (notes.md) in the workspace that tracks what you have tried, what you found and what is left, so that you can pick up the work later. When you have finished, or when you need input, respond without calling any tools.
\end{quote}

A collaboration has three sessions and 15 scripted requester messages. When the prompt exceeds 60,000 tokens, the agent is asked to write a summary of its work, and the history is replaced by the first requester message and this summary. Each new session starts from the system prompt extended by the Markdown documents in the agent's workspace, which serve as project memory. Agents can manage external memory to preserve context across long interactions \citep{packer2023memgpt}.

Table~\ref{tab:workstats} describes the work that precedes a reply. A tool-calling turn is a message of the agent that contains one or more tool calls.

\begin{table}[t]
\centering
\small
\setlength{\tabcolsep}{4pt}
\begin{tabular}{@{}lccc@{}}
\toprule
Model & \shortstack{Tool-calling\\turns} & \shortstack{Reply\\length} & Compactions \\
\midrule
Claude Opus 5 & 5 & 2,645 & 6.25 \\
DeepSeek V4.1 Flash & 4 & 2,136 & 3.75 \\
GLM-5.3 & 2 & 1,710 & 0.25 \\
DeepSeek V4 Pro & 1 & 2,169 & 0.25 \\
Kimi K2.6 & 1 & 2,394 & 0.25 \\
GPT-5.6 Sol & 1 & 2,552 & 0 \\
MiniMax M3 & 1 & 2,232 & 0 \\
Qwen3.6 Plus & 1 & 1,672 & 0 \\
\midrule
All & 2 & 2,143 & 1.34 \\
\bottomrule
\end{tabular}
\caption{Median number of tool-calling turns before a reply, median reply length in characters, and mean number of context compactions per collaboration.}
\label{tab:workstats}
\end{table}

Table~\ref{tab:prevalence} gives the values shown in Figure~\ref{fig:failures}a and~\ref{fig:failures}b.

\begin{table*}[t]
\centering
\small
\setlength{\tabcolsep}{5pt}
\begin{tabular}{@{}lcccccc@{}}
\toprule
 & & & & \multicolumn{3}{c}{Tool-calling turns before the reply} \\
\cmidrule(l){5-7}
Model & Weights & Failure rate & Information gap & 0 & 1--2 & 3 or more \\
\midrule
Claude Opus 5 & closed & 0.82 [0.75, 0.90] & 0.78 & 0.67 & 0.56 & 0.89 \\
DeepSeek V4.1 Flash & open & 0.65 [0.53, 0.77] & 0.63 & 0.71 & 0.50 & 0.71 \\
MiniMax M3 & open & 0.53 [0.37, 0.70] & 0.37 & 0.43 & 0.58 & 0.62 \\
GLM-5.3 & open & 0.52 [0.43, 0.60] & 0.42 & 0.43 & 0.42 & 0.68 \\
DeepSeek V4 Pro & open & 0.50 [0.42, 0.62] & 0.43 & 0.56 & 0.28 & 0.76 \\
Qwen3.6 Plus & open & 0.50 [0.33, 0.67] & 0.42 & 0.63 & 0.37 & 0.50 \\
Kimi K2.6 & open & 0.48 [0.38, 0.57] & 0.45 & 0.39 & 0.44 & 0.63 \\
GPT-5.6 Sol & closed & 0.30 [0.22, 0.42] & 0.20 & 0.20 & 0.24 & 0.54 \\
\midrule
All &  & 0.538 [0.475, 0.600] & 0.46 & 0.48 & 0.40 & 0.70 \\
\bottomrule
\end{tabular}
\caption{Failure rate of the 60 replies of each model with 95\% intervals, share of replies with at least one information gap, and failure rate by the number of tool-calling turns before the reply. The three groups of turns contain 120, 174 and 186 replies.}
\label{tab:prevalence}
\end{table*}

\paragraph{Results by project.}\label{app:projects} Table~\ref{tab:projects} reports the failure rate and the effect of the two interventions separately for the four projects.

\begin{table*}[t]
\centering
\small
\setlength{\tabcolsep}{5pt}
\begin{tabular}{@{}llcccc@{}}
\toprule
 & & & \multicolumn{3}{c}{At the reply points} \\
\cmidrule(l){4-6}
Project & Type & Failure rate & Default & Requirements & Rewrite \\
\midrule
Mislabeled examples & research & 0.66 & 0.73 & 0.53 & 0.38 \\
Long-document QA & research & 0.50 & 0.41 & 0.34 & 0.26 \\
Subscription pricing & business & 0.41 & 0.44 & 0.35 & 0.39 \\
Onboarding redesign & business & 0.58 & 0.50 & 0.47 & 0.38 \\
\bottomrule
\end{tabular}
\caption{Failure rate by project over the 120 replies of each project, and at the reply points by default, under the requirements and after the rewrite.}
\label{tab:projects}
\end{table*}

\section{Writing Conditions}
\label{app:conditions}

The notice reads:
\begin{quote}\small
Note: the user has only seen their own messages and the replies you sent without tool calls. They have not seen your tool calls, tool outputs, or any documents or notes you wrote.
\end{quote}

The requirements read:
\begin{quote}\small
The user has seen only their own messages and the replies you sent without tool calls, not your tool calls, outputs, notes or documents. Write so that they can act on your message without reading anything else: answer their latest message first, say what was done, the result, the next step, and where to check the evidence, and define any name they have not seen.
\end{quote}

In the separate-pass conditions the model receives the system prompt of the working context and one message with the prompt below. In the fact-list condition the record is replaced by a list that the model produces in its working context in response to the instruction to list the results that bear on the user's latest message: findings with numbers, changes made, problems or limitations, and decisions the user needs to make. Collaborations in Chinese use the Chinese versions of these prompts.

\smallskip\noindent\textbf{Separate pass.}\par
{\ttfamily\scriptsize\raggedright\setlength{\parindent}{0pt}\setlength{\parskip}{0pt}\par
Below are your conversation with a user, the files the user provided, and the full record of your work for them (your tool calls and their outputs).\par
\vspace{0.6\baselineskip}
[User\textquotesingle{}s files]\par
\{files\}\par
\vspace{0.6\baselineskip}
[Conversation so far]\par
\{dialog\}\par
\vspace{0.6\baselineskip}
[User\textquotesingle{}s latest message]\par
\{umsg\}\par
\vspace{0.6\baselineskip}
[Work record]\par
\{record\}\par
\vspace{0.6\baselineskip}
Reply to the user\textquotesingle{}s latest message.\par
}

\smallskip\noindent\textbf{Fact list.}\par
{\ttfamily\scriptsize\raggedright\setlength{\parindent}{0pt}\setlength{\parskip}{0pt}\par
Below are your conversation with a user, the files the user provided, and a list of results from your work for them.\par
\vspace{0.6\baselineskip}
[User\textquotesingle{}s files]\par
\{files\}\par
\vspace{0.6\baselineskip}
[Conversation so far]\par
\{dialog\}\par
\vspace{0.6\baselineskip}
[User\textquotesingle{}s latest message]\par
\{umsg\}\par
\vspace{0.6\baselineskip}
[Results of your work]\par
\{record\}\par
\vspace{0.6\baselineskip}
Reply to the user\textquotesingle{}s latest message.\par
}

\smallskip\noindent\textbf{Separate pass with requirements.}\par
{\ttfamily\scriptsize\raggedright\setlength{\parindent}{0pt}\setlength{\parskip}{0pt}\par
Below are your conversation with a user, the files the user provided, and the full record of your work for them (your tool calls and their outputs).\par
\vspace{0.6\baselineskip}
[User\textquotesingle{}s files]\par
\{files\}\par
\vspace{0.6\baselineskip}
[Conversation so far]\par
\{dialog\}\par
\vspace{0.6\baselineskip}
[User\textquotesingle{}s latest message]\par
\{umsg\}\par
\vspace{0.6\baselineskip}
[Work record]\par
\{record\}\par
\vspace{0.6\baselineskip}
Reply to the user\textquotesingle{}s latest message. The user has seen only their own messages and the replies you sent without tool calls, not your tool calls, outputs, notes or documents. Write so that they can act on your message without reading anything else: answer their latest message first, say what was done, the result, the next step, and where to check the evidence, and define any name they have not seen.\par
}

The reply points are selected within each collaboration. Replies written after a context compaction or at the start of a later session are selected first, and the remaining points are drawn at random. Table~\ref{tab:permodel} reports the failure rates by model.

\begin{table*}[t]
\centering
\small
\setlength{\tabcolsep}{4pt}
\begin{tabular}{@{}lcccccc@{}}
\toprule
Model & Default & Notice & Requirements & \shortstack{Separate\\pass} & \shortstack{Fact\\list} & \shortstack{Separate pass\\with requirements} \\
\midrule
Claude Opus 5 & 0.72 & 0.84 & 0.53 & 0.72 & 0.75 & 0.50 \\
DeepSeek V4.1 Flash & 0.66 & 0.44 & 0.53 & 0.69 & 0.59 & 0.50 \\
DeepSeek V4 Pro & 0.47 & 0.44 & 0.34 & 0.28 & 0.38 & 0.34 \\
GLM-5.3 & 0.59 & 0.59 & 0.47 & 0.50 & 0.66 & 0.41 \\
GPT-5.6 Sol & 0.25 & 0.16 & 0.22 & 0.25 & 0.25 & 0.25 \\
Kimi K2.6 & 0.56 & 0.41 & 0.47 & 0.50 & 0.44 & 0.44 \\
MiniMax M3 & 0.53 & 0.53 & 0.59 & 0.53 & 0.47 & 0.59 \\
Qwen3.6 Plus & 0.41 & 0.56 & 0.25 & 0.31 & 0.41 & 0.34 \\
\midrule
All & 0.52 & 0.50 & 0.43 & 0.47 & 0.49 & 0.42 \\
\bottomrule
\end{tabular}
\caption{Failure rate by model under the six writing conditions, on 32 reply points per model.}
\label{tab:permodel}
\end{table*}

\section{Rewriting}
\label{app:rewriting}

The rewriting model receives the request, the work record and the default reply, and it rewrites the reply once with the prompt below. Collaborations in Chinese use the Chinese version of this prompt. Table~\ref{tab:rewrite-family} gives the failure rates of the 256 default replies and their rewrites by family.

\smallskip\noindent\textbf{Rewriting.}\par
{\ttfamily\scriptsize\raggedright\setlength{\parindent}{0pt}\setlength{\parskip}{0pt}\par
You are revising a message that an AI agent is about to send to the person who asked for the work.\par
\vspace{0.6\baselineskip}
[Work record]\par
\{traj\}\par
\vspace{0.6\baselineskip}
[The request]\par
\{req\}\par
\vspace{0.6\baselineskip}
[Current message]\par
\{msg\}\par
\vspace{0.6\baselineskip}
Rewrite the whole message. The user has seen only their own messages and the replies you sent without tool calls, not your tool calls, outputs, notes or documents. Write so that they can act on your message without reading anything else: answer their latest message first, say what was done, the result, the next step, and where to check the evidence, and define any name they have not seen. Use only facts from the work record, keep the correct content of the current message, and write in English. Output only the revised message.\par
}

\begin{table}[t]
\centering
\small
\begin{tabular}{@{}lrcc@{}}
\toprule
Family & $n$ & Default & Rewrite \\
\midrule
Claude & 32 & 0.72 & 0.47 \\
DeepSeek & 64 & 0.56 & 0.38 \\
GLM & 32 & 0.59 & 0.59 \\
GPT & 32 & 0.25 & 0.13 \\
Kimi & 32 & 0.56 & 0.38 \\
MiniMax & 32 & 0.53 & 0.28 \\
Qwen & 32 & 0.41 & 0.25 \\
\bottomrule
\end{tabular}
\caption{Failure rate of the default replies and their rewrites by family.}
\label{tab:rewrite-family}
\end{table}

\paragraph{Human evaluation.}\label{app:humanrewrite} The evaluation covers 83 reply points from ten collaborations of the rewriting experiment. The ten collaborations consist of one per model, drawn so that each of the four projects occurs twice, and two more, one in each language, which gives five collaborations per language. Each point has two versions, the default reply and its rewrite, which share the requester's files and the conversation so far. Five annotators follow the protocol of Appendix~\ref{app:agreement}. Each annotator sees one version per point and is not told that versions exist, three annotators see one version and two the other, and the assignment alternates over the points, which gives 415 annotations.

The measure is the share of the annotators of a version who label the reply as failed, and the contrast is the difference between the default reply and the rewrite at the same point, averaged over the points, with an interval from a bootstrap over the collaborations. The share is 0.52 for the default replies and 0.32 for the rewrites, a difference of 0.201 [0.095, 0.321]. The share is lower for the rewrite in seven of the eight models. Annotators who see the same reply agree at a mean pairwise $\kappa$ of 0.60. On these points the reader fails 42 of the 83 default replies and 31 of the 83 rewrites, rates of 0.51 and 0.37. With the majority label of the annotators of a reply, the reader agrees at $\kappa = 0.60$ over the 166 replies, with an accuracy of 80\%. The reader fails the default reply and passes the rewrite at 21 points, and at 17 of them the majority labels are the same pair.

\section{Gap Categories}
\label{app:gaps}

The classifier is Gemini 3 Flash (preview) with thinking disabled and temperature 0. Each call receives one gap: the reader's question, the span of the reply, the type assigned by the reader, and 200 characters of the reply on either side of the span. The prompt gives the category definitions of Section~\ref{sec:gaps}. It states that entries of the literature index, file names, sections of documents and outputs of runs are pointers to unseen material, and it restricts the term category to abbreviations and terms that are established in the field. Table~\ref{tab:gapfamily} gives the shares by family. Of the 480 replies, 258 have no information gap, 108 have one, 49 have two and 65 have three or more.

Two annotators independently labeled 100 of the 475 gaps, sampled in proportion to the number of gaps of each model. They received what the classifier receives, with the same category definitions. The classifier agrees with the two annotators at Cohen's $\kappa$ of 0.69 and 0.67, and the annotators agree with each other at $\kappa = 0.68$. On the gaps to which both annotators assign the same category, the classifier agrees with that category at $\kappa = 0.82$.

Figure~\ref{fig:failures}c covers the 198 coined labels. To locate the earliest occurrence of a coined label, we search the messages of the agent's context in order for the first message that contains a term of its span, ignoring case and markup. The message type gives the source, and a label without an earlier occurrence is counted as first used in the reply itself. A first occurrence in a system message is an occurrence in the project memory, which the environment loads at the start of the second and third session.

\begin{table*}[t]
\centering
\small
\setlength{\tabcolsep}{4pt}
\begin{tabular}{@{}lrcccccc@{}}
\toprule
Family & \shortstack{Failed\\replies} & \shortstack{Pointer to\\unseen material} & \shortstack{Coined\\label} & \shortstack{Unexplained\\term} & \shortstack{Missing\\background} & \shortstack{Request not\\answered} & \shortstack{Wrong\\language} \\
\midrule
Claude & 49 & 61.2 & 65.3 & 12.2 & 0.0 & 18.4 & 8.2 \\
DeepSeek & 69 & 44.9 & 49.3 & 33.3 & 0.0 & 27.5 & 0.0 \\
GLM & 31 & 48.4 & 45.2 & 16.1 & 0.0 & 35.5 & 0.0 \\
GPT & 18 & 16.7 & 0.0 & 44.4 & 11.1 & 38.9 & 0.0 \\
Kimi & 29 & 51.7 & 37.9 & 24.1 & 3.4 & 20.7 & 6.9 \\
MiniMax & 32 & 34.4 & 31.2 & 12.5 & 0.0 & 43.8 & 15.6 \\
Qwen & 30 & 46.7 & 26.7 & 26.7 & 3.3 & 23.3 & 20.0 \\
\midrule
All & 258 & 46.1 & 42.2 & 23.6 & 1.6 & 28.3 & 6.6 \\
\bottomrule
\end{tabular}
\caption{Share of failed replies (\%) with each kind of failure, by family.}
\label{tab:gapfamily}
\end{table*}

\section{Sampling Analysis}
\label{app:sampling}

The 100 reply points are drawn at random from the reply points of collaborations that Qwen3.6-35B-A3B ran as the agent. The model runs locally with llama.cpp, with thinking disabled. With $c$ usable replies among the $n$ samples of a point, the probability that $k$ samples drawn without replacement contain at least one usable reply is $1 - \binom{n-c}{k} / \binom{n}{k}$, and pass@$k$ \citep{chen2021evaluating} averages this quantity over the points. \citet{brown2024large} study the same quantity for task success under repeated sampling.

\end{document}